%% file: bare_jrnl_new_sample4.tex
\documentclass[lettersize,journal]{IEEEtran}
\usepackage{amsmath,amsfonts}
\usepackage{algorithmic}
\usepackage{algorithm}
\usepackage{array}
\usepackage[caption=false,font=normalsize,labelfont=sf,textfont=sf]{subfig}
\usepackage{textcomp}
\usepackage{stfloats}
\usepackage{url}
\usepackage{verbatim}
\usepackage{graphicx}
\usepackage{cite}
\usepackage{amssymb}
\usepackage{subfiles}
\usepackage{multirow}
\usepackage{booktabs}
\usepackage{xcolor}
\usepackage{colortbl}

\begin{document}

\title{MC-GPT: Empowering Vision-and-Language Navigation with Memory Map and Reasoning Chains}

\author{Zhaohuan Zhan, Lisha Yu, Sijie Yu, Guang Tan
\thanks{Zhaohuan Zhan, Sijie Yu and Guang Tan are with the School of Intelligent Systems Engineering, Shenzhen Campus of Sun Yat-sen University, Shenzhen, Guangdong, 518107, P.R.China. (Corresponding author: Guang Tan.) (e-mail: zhanzhh5@mail2.sysu.edu.cn; yusj8@mail2.sysu.edu.cn; tanguang@mail.sysu.edu.cn)

Lisha Yu is with the School of Computer Science and Engineering, Guangzhou, Guangdong, 510006, P.R.China. (e-mail:yulsh6@mail2.sysu.edu.cn)}}

\maketitle

\begin{abstract}
In the Vision-and-Language Navigation (VLN) task, the agent is required to navigate to a destination following a natural language instruction. While learning-based approaches have been a major solution to the task, they suffer from high training costs and lack of interpretability. Recently, Large Language Models (LLMs) have emerged as a promising tool for VLN due to their strong generalization capabilities. However, existing LLM-based methods face limitations in memory construction and diversity of navigation strategies.
  
To address these challenges, we propose a suite of techniques. Firstly, we introduce a method to maintain a topological map that stores navigation history, retaining information about viewpoints, objects, and their spatial relationships. This map also serves as a global action space. Additionally, we present a Navigation Chain of Thoughts module, leveraging human navigation examples to enrich navigation strategy diversity. Finally, we establish a pipeline that integrates navigational memory and strategies with perception and action prediction modules. Experimental results on the REVERIE and R2R datasets show that our method effectively enhances the navigation ability of the LLM and improves the interpretability of navigation reasoning.
\end{abstract}


\begin{figure*}[htb]
    \centering
    \includegraphics[width=1.0\textwidth]{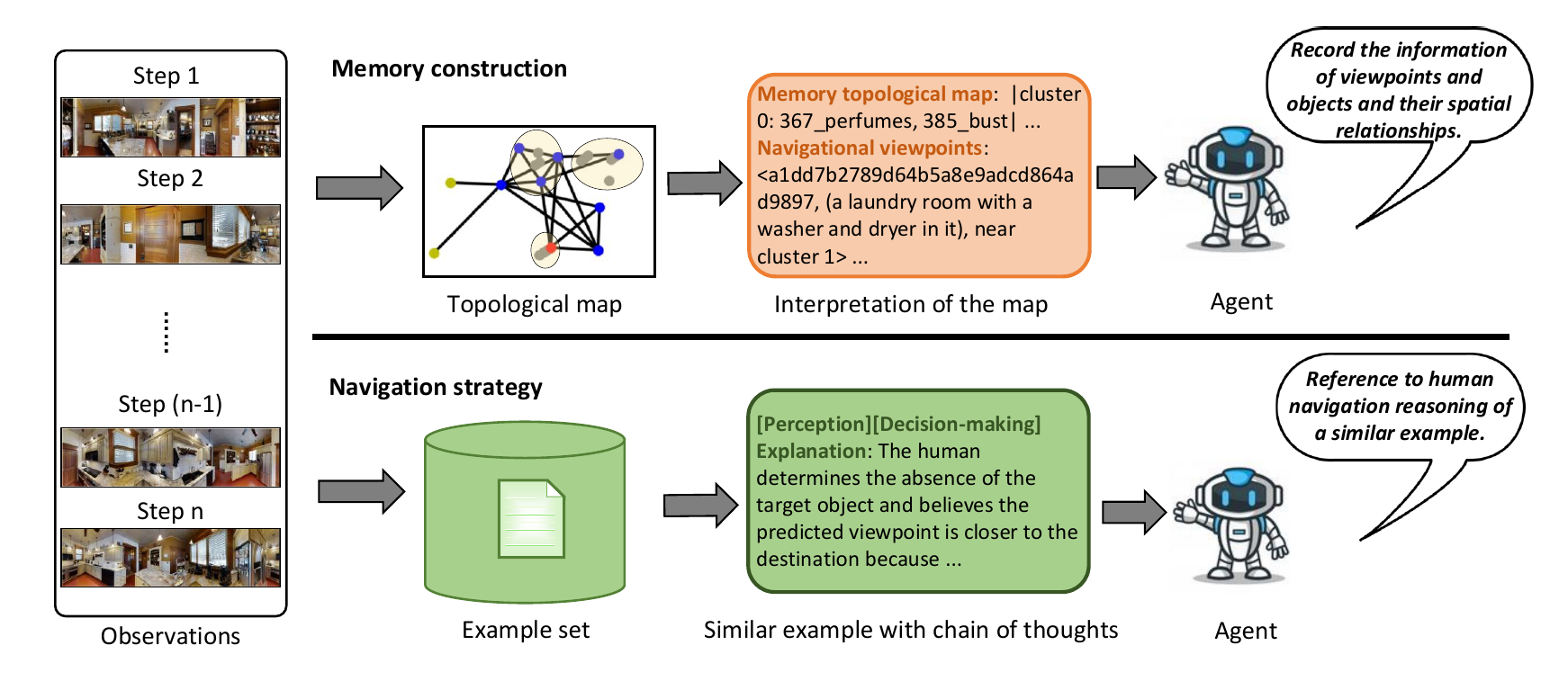}
    \caption{Accurate memory and suitable strategies are crucial for navigation tasks. A topological map can precisely record information about viewpoints, while incorporating human navigation reasoning examples can enhance the diversity of strategies employed by the LLM.}
  \label{fig:intuition}
\end{figure*}

\subfile{sections/introduction}
\subfile{sections/related}
\subfile{sections/approach}
\subfile{sections/setup}
\subfile{sections/result}
\subfile{sections/conclusion}

\vfill

\end{document}

%% file: sections/introduction.tex
\section{Introduction} \label{intro}
\IEEEPARstart{I}{n} recent years, the convergence of computer vision and natural language processing has given rise to an emerging task known as Vision-and-Language Navigation (VLN) \cite{anderson2018vision, qi2020reverie, ku2020room}. This cross-modal task necessitates the integration of visual perception and natural language understanding, enabling autonomous agents to understand their surrounding environments and navigate to destinations. The VLN task requires an agent to follow a natural language instruction and navigate to the destination.

The current mainstream approach to solving the VLN task is learning-based methods. By training on VLN datasets, the navigation policy is implicitly defined by the network parameters. However, these methods have notable drawbacks: on one hand, they incur high training costs and are prone to overfitting the training set; on the other hand, the inference process lacks interpretability and fails to incorporate common sense knowledge.

Recently, Large Language Models (LLMs) \cite{brown2020language,driess2023palm,touvron2023llama} have emerged as a promising approach to addressing the issues with VLN tasks. The generalization ability of LLMs can help mitigate overfitting issues and reduce training costs, while the common sense knowledge of LLMs can provide enhanced interpretability for navigation reasoning. Some efforts have integrated LLMs as sub-modules within traditional navigation frameworks, allowing LLMs to handle subtasks such as instruction decomposition and landmark extraction \cite{shah2023lm,zhou2023esc,qiao2023march, cai2023bridging}. We refer to this as the \textbf{LLM-sub} approach. With this approach, the overfitting issue persists as it still requires training. Conversely, a few studies have positioned LLMs as core controllers in navigation decision-making~\cite{zhou2023navgpt,chen2024mapgpt, dorbala2023can,schumann2023velma} without any additional training, an approach we call \textbf{LLM-core}. In this paper, we adopt the LLM-core approach in order to fully exploit the potential of the LLM and maximize the generalization ability of our method.  

Despite the advancements, existing LLM-core methods tend to oversimplify problem modeling. For instance, most methods utilize a combination of Vision-and-Language Model (VLM)~\cite{li-etal-2023-lavis,liu2024llava,achiam2023gpt4, yang2023effective} and LLM to summarize the navigation history as a context, or memory, for LLM interaction. This approach not only introduces noise but also requires multiple calls of models, which increases time consumption and API costs. It is suggested that the key information that affects VLN performance includes \textbf{objects}, \textbf{room type}, and \textbf{orientation}~\cite{qi2021know,gao2021room}. Therefore, we propose constructing a memory structure that accurately captures the navigation context, and highlights these components. As depicted in Fig. \ref{fig:intuition}, we introduce a topological map as the memory structure. The map records the information about viewpoints and objects, as well as their spatial relationships. During navigation, the agent adds observations as new nodes to the topological map, and updates the layout of objects accordingly. The connectivity between nodes preserves the \textbf{orientation} information, while the object layouts retains \textbf{object} information, and allows for inferring \textbf{room-type} information. 

Typically, the current LLM-core methods need to provide basic rules for navigation, a technique known as {\em Standard Prompting}~\cite{brown2020language}. However, prescribing the reasoning process can limit the agent's navigation strategies, and has difficulty incorporating domain-specific knowledge. In reality, humans may employ different navigation strategies for different scenarios. For example, executing the instructions \emph{``go to the kitchen to get some milk''} and \emph{``go to the utility room to get a stool''} requires distinct navigation strategies. Here, ``kitchen'' is an easily recognizable room type, which provides clear directional guidance. In contrast, the utility room may be less common, with a somewhat ambiguous definition, as it may coexist with other areas of a house, such as the laundry room. In essence, executing the first task may involve a simple room-type based priority strategy, while the second task requires a strategy that balances room-type priority with object priority, alongside commonsense reasoning.

The reasoning process on such a case-by-case basis can be addressed by having the LLM look at demonstration examples~\cite{brown2020language,min2022rethinking}. The benefit is more pronounced when the offered examples contain a step-by-step reasoning process, a method called the Chain-of-Thought (CoT) prompting \cite{wei2022chain,zhang2022automatic}. Inspired by these techniques, we introduce a {\em Navigation Chain of Thought Module}, as depicted in Fig. \ref{fig:intuition}. We first establish an example set covering various scenarios, and employ an LLM to mine the reasoning chains of these examples; during testing, the LLM can retrieve a demonstration example from this set. Notably, the examples are collected from navigation datasets such as REVERIE \cite{qi2020reverie}. These datasets are annotated by humans and thus reflect human navigation decision-making. Our approach of navigation does not need extra manual annotation.

In summary, the contributions of this paper can be summarized as follows:
\textbf{1)} We propose the use of a memory topological map to store key navigation information. This map retains rich spatial-temporal information and is suitable for object-oriented navigation tasks, while addressing the drawbacks of current methods in memory modeling. \textbf{2)} We propose a mechanism called Navigation Chain of Thought, which can provide a demonstration example with a chain of thoughts to the LLM. The module can enable the LLM to employ case-by-case navigation strategies and thus increase the diversity of navigation strategies. \textbf{3)} We present a pipeline to manage the navigation process. This pipeline includes perception, memory, commonsense, and decision-making modules, allowing for fine decoupling and modeling of navigation problems. \textbf{4)} Experimental results on the REVERIE and R2R datasets demonstrate that our approach improves navigation performance, enhancing the interpretability of navigation reasoning, and providing new ideas and tools for bridging the gap between human and robotic navigation.





%% file: sections/related.tex
\section{Related work}
\subsection{Vision-and-Language Navigation}
The Vision-and-Language Navigation (VLN) task~\cite{anderson2018vision,qi2020reverie, ku2020room}  requires robots to understand natural language navigation instructions and move to target locations by executing a series of navigation actions. Current VLN methods primarily focus on improving navigation success rates and efficiency by optimizing model structure \cite{anderson2018vision,hong2021vln,chen2022think,qiao2022hop,liang2022visual,lin2022adapt,an2022bevbert}, representation learning\cite{wang2020soft,wang2019reinforced}, and data augmentation\cite{fu2020counterfactual, zhan2023object}.

To address the issue of data scarcity, Fu \emph{et al.} \cite{fu2020counterfactual} utilize counterfactual reasoning for data augmentation. They make the path sampler and navigator against each other during training based on the principles of adversarial learning. Wang \emph{et al.} \cite{wang2020soft} improve supervised signals by introducing additional terms to the loss function of the reinforcement learning based on knowledge distillation, making the supervision signals softer and more robust. Wang \emph{et al.} \cite{wang2019reinforced} find that training the navigator with the shortest path as the sole supervision signal may lead to biases in training when human annotations do not correspond strictly with the shortest path. Therefore, using the distance to the navigation endpoint as an additional supervision signal can enhance training robustness. This training strategy can also be generalized to synthetic data. With the rise of pre-trained models and the development of prompting engineering, a series of VLN methods \cite{hong2021vln,chen2022think,qiao2022hop,liang2022visual,lin2022adapt,an2022bevbert} based on pre-trained models have been proposed. These methods improve the effectiveness of VLN algorithms in various ways, although they may result in increased costs in computation and training time.

\subsection{Navigation with LLM}

LLM's generalization and reasoning capabilities have sparked interest in navigation tasks \cite{brown2020language,driess2023palm, touvron2023llama, zhou2024navgpt}. Some researchers attempt to use the LLM as an auxiliary module in the traditional navigation frameworks, in which the LLM provides navigation guidance or helps with landmark extraction. Shah \emph{et al.} \cite{shah2023lm} utilize the GPT-3 to identify landmarks in instructions, aligning textual and visual modalities using a VLM (CLIP \cite{radford2021learning}), and then integrate them into a traditional navigation framework to complete navigation reasoning. Zhou \emph{et al.} \cite{zhou2023esc} introduce VLM as a scene understanding module, leveraging large language models to provide commonsense knowledge, using probabilistic soft logic to formulate reasoning rules, and finally combining them with frontier exploration strategies to predict navigation actions. Qiao \emph{et al.} \cite{qiao2023march} focus on high-level concise navigation instructions, employing large language models to generate fine-grained instructions. They predict target objects, destinations, and current objects and room types, and when a change in room type is perceived, they use the LLM to generate fine-grained next-step instruction. Cai \emph{et al.} \cite{cai2023bridging} utilize the GPT-4 \cite{achiam2023gpt4} for self-localization and next-step decision-making, then use a pixel-based navigation algorithm to complete subsequent navigation actions. 

In other works, LLM is employed as the core controller of navigation, where the LLM directly makes navigation decisions and interacts with other modules within the navigation system. Zhou \emph{et al.} \cite{zhou2023navgpt} describe visual perception using VLM, detect objects using an object detector, and then employ the ChatGPT to summarize historical information and make specific reasoning. Chen \emph{et al.} \cite{chen2024mapgpt} introduce a global map into an LLM-based navigator, expanding the navigator's action space from local to global. Dorbala \emph{et al.} \cite{dorbala2023can} use the LLM for navigation decision-making, directing navigation towards the target direction when no target objects are perceived, and navigating around the target object when it appears in sight. Schumann \emph{et al.} \cite{schumann2023velma} propose an embodied LLM agent, using language descriptions of trajectories and visual environment observations as context hints for the next action, and outputting navigation actions based on LLM's next-word prediction. Different from previous works, our method aims to address the limitations of LLM-based navigators in memory construction and the diversity of navigation strategies. We also propose a novel pipeline for LLM-based navigation.

%% file: sections/approach.tex
\section{Approach}
\subsection{Problem setup}
According to the setting of the VLN task, an agent is randomly placed at a viewpoint and receives an instruction. To carry out the instruction, the agent needs to reason step-by-step by grounding the textual instruction and visual observation. At each time step, the agent selects and moves to one of the navigable viewpoints according to the instruction and the observation. Under the REVERIE task (object-oriented VLN task) setting, the agent also needs to carry out the referring expression task \cite{qiao2020referring} - point out the referred object from a set of object proposals.


Formally, let $\emph{{I}}$ be the navigation instruction, and we can obtain the visual observation by calling the function of MP3D simulator \cite{anderson2018vision}:

\begin{equation}
    {V_t}{=\left\{ {{\emph{{v}}_{j}}} \right\}_{j = 1}^J=\rm{get\_observation()}}
\end{equation}
where \emph{J} is the number of discrete views. Among \emph{J} discrete views, only \emph{L} views could be reached, we call them local navigable viewpoints and denote them as $\hat{\emph{{{V}}}}=\left\{ {\emph{{v}}_{l}} \right\}_{l=1}^L$, where $L \leq J$. By combining the previous local navigable viewpoints and a ``STOP" state, we can formulate the global navigational viewpoints a.k.a action space: 
\begin{equation}
\begin{aligned}
    {\emph{{U}}_t} = \left\{ {{\emph{{u}}_{n}}} \right\}_{n = 1}^N = \sum_{h=0}^t \hat{\emph{{{V}}}}_h \cup \left\{STOP\right\}\\
    =\rm{get\_navigational\_viewpoints}(\emph{V}_t)
\end{aligned}
\end{equation}
where $N$ is the number of global navigable viewpoints. At each time step, the agent selects and moves to one of the navigable viewpoints $ {\emph{{U}}_t}$. Under the object-oriented navigation task setting such as the REVERIE task, the agent also needs to point out the referred object from a set of object proposals:
\begin{equation}
    \emph{{O}}_t = \left\{ {{\emph{{o}}_k}} \right\}_{k = 1}^K = \rm{get\_object\_proposals}(\emph{V}_t)
\end{equation}
where  \emph{K} is the number of object proposals.

\subsection{Overview}
\begin{figure*}[h]
    \centering
    \includegraphics[width=1.0\textwidth]{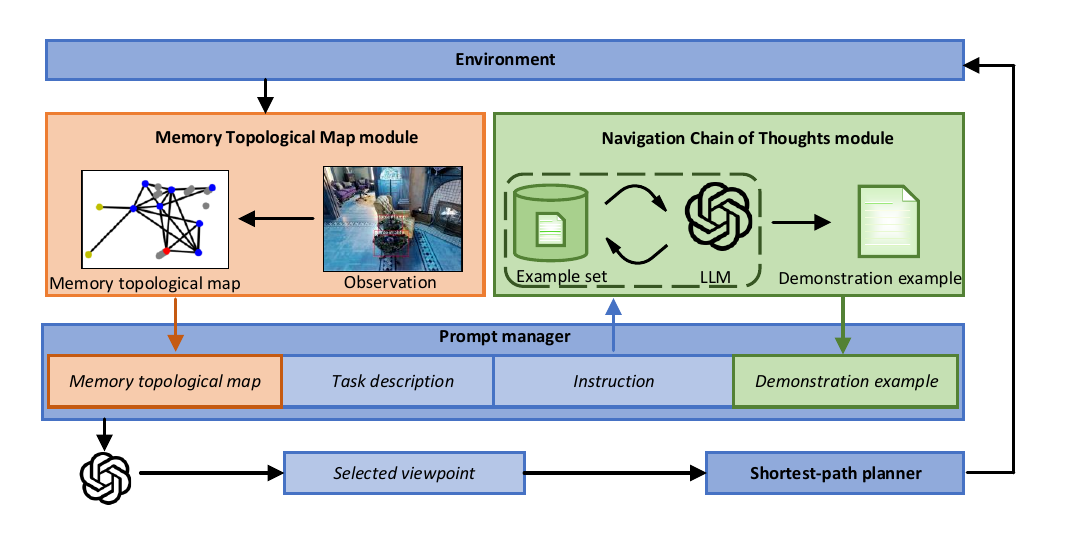}
    \caption{Overview of our pipeline. The Memory Topological Map module manages a map for storing memory, while the Navigation Chain of Thoughts module provides a demonstration example with a reasoning chain. The Prompt manager organizes all essential information and feeds it to the LLM. The LLM then makes a viewpoint selection and employs a Shortest-path Planner to navigate to the selected viewpoint.
    }
    \label{fig: pipeline}
\end{figure*}

Fig. \ref{fig: pipeline} and Algorithm \ref{alg1} present our pipeline. Initially, our agent acquires visual observations of the surrounding environments and stores them in memory using the \emph{\textbf{Memory Topological Map module}}. Further elaboration on the Memory Topological Map module can be found in Section \ref{map}. Subsequently, our agent utilizes the \emph{\textbf{Navigation Chain of Thoughts module}} to query an example with a chain of thoughts by computing the similarities between the test instance and the example set. More details regarding the Navigation Chain of Thoughts module are provided in Section \ref{cot}. Finally, the agent organizes the information, including Task Description, Instruction, Memory Topological Map, and Demonstration Example, using the \emph{\textbf{Prompt Manager}}, and passes them to the LLM. Details of the Prompt Manager can be found in Section \ref{prompt}. The LLM then conducts reasoning and decides which viewpoint to select. Subsequently, the agent employs a Shortest-Path Planner to move to the selected viewpoint.

\begin{algorithm}
\caption{Task Execution Algorithm}
\label{alg1}
\begin{algorithmic}[1]

\STATE \textbf{Input:} Task description $T$, instruction $I$, example set $D = \{d_1, d_2, \ldots, d_Q\}$, maximum number of navigation steps $max\_episode\_length$

\STATE \textbf{Output:} Selected viewpoint $u_n$, selected object $o_k$

\STATE \textbf{Operator functions:}
\STATE $\rm{F_M}$: Memory Topological Map module
\STATE $\rm{F_C}$: Navigation Chain of Thoughts module
\STATE $\rm{F_P}$: Prompt manager
\STATE $\rm{LLM}$: Large language model
\STATE $\rm{F_S}$: Shortest-path planner

\STATE $step \leftarrow 0$

\FOR{$step < max\_episode\_length$}
    
    \STATE \textcolor{cyan!50!blue}{// Obtain observation, navigational viewpoints and object proposals from simulator}
    \STATE $V_t \leftarrow get\_observation()$
    \STATE $U_t \leftarrow get\_navigational\_viewpoints(V_t)$
    \STATE $O_t \leftarrow get\_object\_proposals(V_t)$

    \STATE \textcolor{cyan!50!blue}{// Construct memory topological map}
    \STATE $M_t \leftarrow \rm{F_M}(\emph{U}_t, \emph{O}_t, \emph{M}_{t-1})$
    
    \STATE \textcolor{cyan!50!blue}{// Query a chain of thought example}
    \STATE $d_q \leftarrow \rm{F_C}(\emph{I}, \emph{D})$
    
    \STATE \textcolor{cyan!50!blue}{// Build prompt}
    \STATE $P_t \leftarrow \rm{F_P}(\emph{M}_t, \emph{T}, \emph{I}, \emph{d}_q)$
    
    \STATE \textcolor{cyan!50!blue}{// Select a navigational viewpoint and object}
    \STATE $(u_n \in U_t, o_k \in O_t) \leftarrow \rm{LLM}(\emph{P}_t)$
    
    \STATE \textcolor{cyan!50!blue}{// Execute action}
    \STATE $\rm{F_S}(\emph{u}_n, \emph{M}_t)$
    
    \IF{$u_n == \rm{STOP}$}
        \STATE \textbf{break}
    \ENDIF
    
    \STATE $step \leftarrow step + 1$
    
\ENDFOR
\RETURN $u_n, o_k$
\end{algorithmic}
\end{algorithm}

\subsection{Memory Topological Map module} \label{map}
\begin{figure*}[h]
    \centering
    \includegraphics[width=1.0\textwidth]{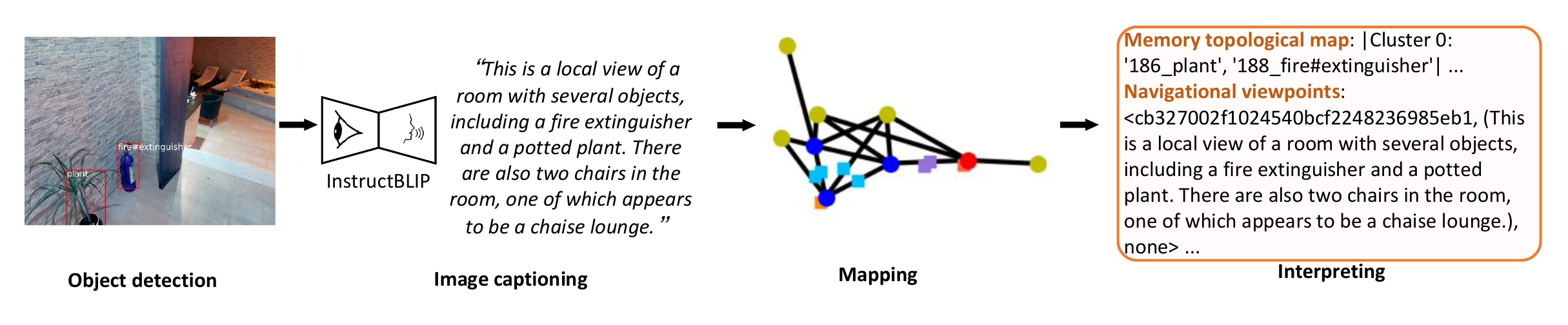}
    \caption{The construction of the Memory Topological Map. Initially, with object detection and image captioning, we can convert visual perceptions into textual descriptions. Subsequently, we construct a map comprising viewpoints and objects. Notably, the {\color{red} red node} represents the viewpoint where the agent is currently located, while the {\color{yellow} yellow nodes} denote unvisited viewpoints, and the {\color{blue} blue nodes} mean visited viewpoints. The squares on the map represent objects, and squares with the same colors indicate that they are clustered together. Finally, we utilize a textual prompt template to interpret the map.}
    \label{fig: map}
\end{figure*}

The Memory Topological Map records the information of observed viewpoints (including visited and unvisited ones) in a topological graph. Formally, the construction of a memory topological map can be represented as:
\begin{equation}
    M_t = \rm{F_M}(\emph{U}_t, \emph{O}_t, \emph{M}_{t-1})
\end{equation}
The operations of $\rm{F_M}$ are illustrated in Fig. \ref{fig: map} and can be roughly divided into five steps as follows.

\textbf{Graph Construction.} We establish and manage a topological graph where observed viewpoints serve as nodes. At each time step, newly observed navigable viewpoints are incorporated as new nodes. If two nodes can navigate to each other, an edge is established between them. This iterative process enables the incremental construction of a topological graph. Notably, this graph also functions as the action space as it contains routes to all observed viewpoints.

\textbf{Object Detection.} For each viewpoint, we retain its object detection results. Specifically, for newly observed viewpoints, we capture images from the current viewpoint and send them to the Faster R-CNN with an attention mechanism \cite{anderson2018bottom} for object detection. For existing viewpoints, we persist with the previously obtained object detection results.

\textbf{Image Captioning.} To enhance comprehension of the visual content within viewpoints, we utilize a VLM named InstructBLIP \cite{li-etal-2023-lavis} to generate captions for their images. To minimize computational costs, we selectively invoke InstructBLIP for neighboring viewpoints only, while retaining previously generated captions for other viewpoints. Some qualitative examples can be found in Sec. \ref{blip}.

\textbf{Mapping.} We then create a topological map incorporating object distribution. Objects detected solely in one viewpoint are positioned at that respective viewpoint. However, for objects observed across multiple viewpoints, we position them at the geometric center of the corresponding viewpoints. Besides, considering the co-occurrence of objects provides insight into the room type, we employ the K-Means clustering algorithm \cite{macqueen1967some,hartigan1979algorithm} to cluster nearby objects together. The resulting object-clustered outcomes are utilized for room-type prediction, which is a sub-task of our navigation reasoning. The reason for the sub-task is that the room-type priority strategy is an important navigation strategy for those recognizable room types.

\textbf{Interpreting.} In order to facilitate the understanding of the map by the LLM, textual interpretation is necessary. Drawing inspiration from \cite{vemprala2023chatgpt}, we adopt a structured language template: ``Memory topological map: $|$cluster ID: \{objects\}$|$, Navigable viewpoints: $<$viewpoint ID, image captioning, relationship with Object clusters$>$". As shown in the top-right of Fig \ref{fig: prompt}, the Memory topological map field provides the object information and the clusters, while the Navigable viewpoints field provides the connection relationship between the viewpoints and clusters. Following interpretation, the map is converted into a textual format, which can be directly provided to the LLM as part of the prompt.

\begin{algorithm}
\caption{CoT Mining Algorithm}
\label{alg2}
\begin{algorithmic}[1]

\STATE \textbf{Input:} Task description $T'$, instruction $I$, sampled training set $S$

\STATE \textbf{Output:} Example set $D$

\STATE \textbf{Operator functions:}
\STATE $\rm{F_M}$: Memory Topological Map module
\STATE $\rm{F_P}$: Prompt manager
\STATE $\rm{LLM}$: Large language model
\STATE $\rm{F_S}$: Shortest-path planner

\STATE \textcolor{cyan!50!blue}{// Initialize an example set}
\STATE $D \gets []$

\FOR{$(i, ground\_truth\_path) \in\ S$}

\STATE \textcolor{cyan!50!blue}{// Initialize a CoT example}
\STATE $d \gets []$

\FOR{$(u_n, o_k) \in\ ground\_truth\_path$}
    
    \STATE \textcolor{cyan!50!blue}{// Obtain observation, navigational viewpoints and object proposals from simulator}
    \STATE $V_t \leftarrow get\_observation()$
    \STATE $U_t \leftarrow get\_navigational\_viewpoints(V_t)$
    \STATE $O_t \leftarrow get\_object\_proposals(V_t)$

    \STATE \textcolor{cyan!50!blue}{// Construct memory topological map}
    \STATE $M_t \leftarrow \rm{F_M}(\emph{U}_t, \emph{O}_t, \emph{M}_{t-1})$

    \STATE \textcolor{cyan!50!blue}{// Build prompt}
    \STATE $P_t \leftarrow \rm{F_P}(\emph{M}_t, \emph{T'}, \emph{I}, \emph{u}_n, \emph{o}_k)$
    
    \STATE \textcolor{cyan!50!blue}{// Explain the reason behind viewpoint and object selection}
    \STATE $explanation \leftarrow \rm{LLM}(\emph{P}_t)$

    \STATE \textcolor{cyan!50!blue}{// Update the CoT example with a ``instruction-path-explanation" pair}
    \STATE Add $(I, u_n, o_k, explanation)$ to $d$
    
    \STATE \textcolor{cyan!50!blue}{// Execute action}
    \STATE $\rm{F_S}(\emph{u}_n, \emph{M}_t)$
    \IF{$u_n == \rm{STOP}$}
        \STATE \textbf{break}
    \ENDIF
    
\ENDFOR

\STATE \textcolor{cyan!50!blue}{// Update the example set}
\STATE Add $d$ to $D$

\ENDFOR
\RETURN $D$
\end{algorithmic}
\end{algorithm}

\subsection{Navigation Chain of Thoughts module} \label{cot}
Inspired by the ICL \cite{brown2020language,min2022rethinking} and CoT prompting \cite{wei2022chain,zhang2022automatic}, we expect that allowing LLM to reference human navigation examples in analogous scenarios can enrich the diversity of navigation strategies, thereby more effectively realizing ``case-by-case analysis'' in navigation reasoning. By querying a similar example with a chain of thoughts, we can offer a demonstration for the LLM. 
\begin{figure*}[h]
    \centering
    \includegraphics[width=1.0\textwidth]{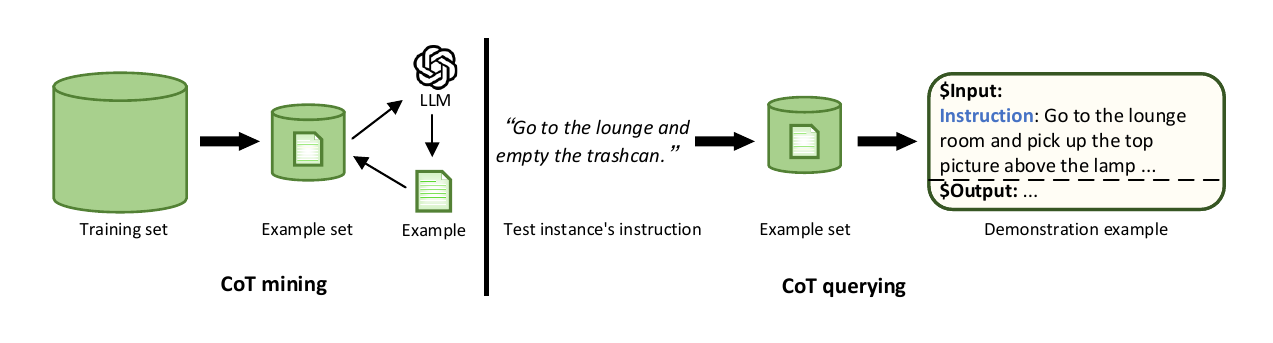}
    \caption{Two steps in the Navigation Chain of Thoughts: CoT mining and CoT querying. CoT mining aims to construct an example set comprising diverse reasoning chains, while CoT querying aims to select a similar example with an appropriate reasoning chain for the LLM's navigation. }
    \label{fig: cot}
\end{figure*}

The training set of the VLN task comprises ``instruction-path'' pairs, where annotators record viewpoint selections while executing the instructions. This training set inherently encompasses diverse examples with varying navigation strategies. As mentioned in Section \ref{intro}, instructions leading to similar destinations often necessitate similar navigation strategies. To distinguish different navigation strategies of the training set, we utilize the destination's room type as a criterion and create a subset of the training set $S$. Specifically, for each ``instruction-path'' pair, we extract the ground-truth room type of the destination from the MatterPort3D simulator \cite{anderson2018vision}. Among the pairs of the training set, there exist a total of 27 room types. We select one example per room type to compose the sampled training set $S$.



\textbf{CoT Mining.} Since $S$ solely consists of instructions and corresponding viewpoint selections (``instruction-path''), this may not provide sufficient insight for the LLM to comprehend the reasoning behind these selections. Due to the unavailability of annotator explanations and the high cost associated with human annotation, we employ an LLM to emulate annotator behavior and mine the reasoning chain of each example. The pseudo-code of CoT mining is presented in Algorithm \ref{alg2}. Most of the processes are similar to the Algorithm \ref{alg1}, except the prompt since we expect the LLM to explain the viewpoint selection rather than predict the viewpoint. For the prompt $P_t$, as shown in the left side of Fig. \ref{fig: cot}, we utilize the role prompting technique \cite{john2023art} to instruct the LLM to play the role of the annotator: we present the LLM with perception and decision-making aspects, prompting it to explain the reason behind each viewpoint selection step by step. More details of our prompt $P_t$ can be found in Sec. \ref{prompt}. After the CoT mining, we extend sample set $S$ from ``instruction-path'' to ``instruction-path-explanation,'' thereby establishing an example set $D$ with chains of thoughts. 

\textbf{CoT Querying.} Ideally, we aim to query an example with the same room type of destination as the test instance. However, determining the room type of the instance's destination requires additional reasoning. To streamline this process, we use a parser to extract the room-type nouns from the instance's instruction. We consider the first room-type noun as the destination's room type based on idiomatic expressions (e.g., ``go to the \emph{kitchen} to get milk''). Subsequently, as shown in the right side of Fig. \ref{fig: cot}, we use the Sentence-BERT \cite{reimers2019sentence} to encode the room-type nouns of the instance and those from the example set. The example with the closest cosine distance to the instance is selected as the demonstration example. In cases where the instruction lacks room-type nouns, we directly compute the cosine distance between the instance's instruction and the room-type nouns from the example set. Then the selected demonstration example with the chain of thoughts is directly integrated into the prompt for the LLM.

Here we represent the CoT querying mathematically. Let the example set be denoted as $D = {d_1, d_2, \ldots, d_Q}$, where $Q$ is the total number of examples. We calculate the cosine similarity between the instruction $I$ and each example $d_i$, then we select the example $d_q$ with the highest cosine similarity to $I$:
\begin{equation}
    d_q = \rm{F_C}(\emph{I}, \emph{D})=\arg\max_{\emph{d}_i \in \emph{D}} \left(\frac{\emph{I} \cdot \emph{d}_i}{\|\emph{I}\| \|\emph{d}_i\|}\right)
\end{equation}
where $\|\|$ denotes the magnitude operation (Euclidean norms) and $\cdot$ represents dot product operation.



\subsection{Prompt manager} \label{prompt}
\begin{figure*}[h]
    \centering
    \includegraphics[width=1.0\textwidth]{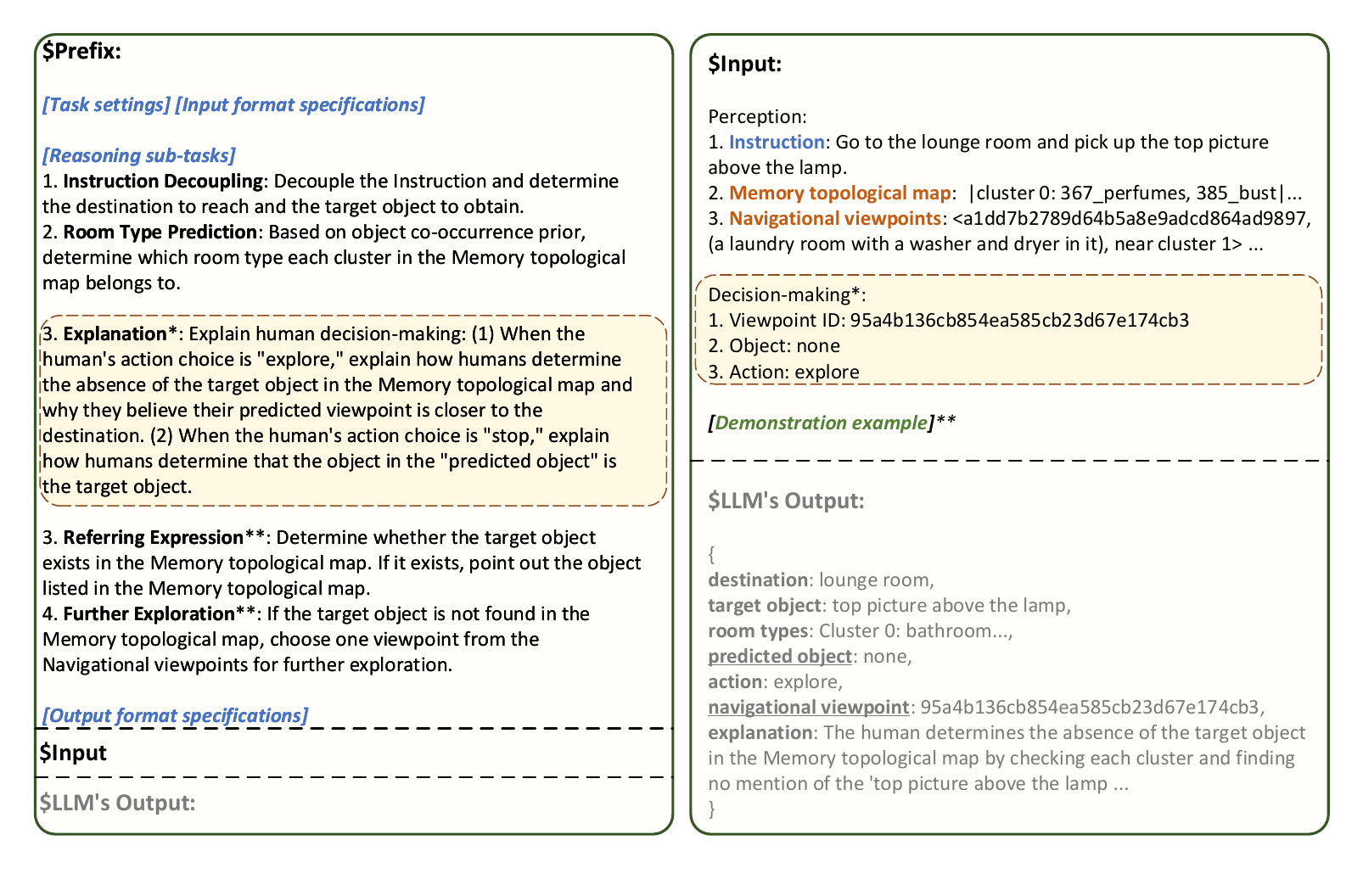}
    \caption{Our prompt manager integrates two types of prompts, the \$\textbf{Prefix} prompt and the \$\textbf{Input} prompt. The former is designed to illustrate basic reasoning rules, while the latter provides perceptual information. The prompt manager is employed for both CoT mining and navigation inference. Thus, the symbol $\ast$ denotes content exclusively relevant during CoT mining, whereas $\ast\ast$ means content exclusive to navigation inference.}
    \label{fig: prompt}
\end{figure*}

To aid the LLM in understanding the agent's perception and making the appropriate decisions, we introduce a prompt manager $F_P$ designed to organize all kinds of information. The prompt is depicted in Fig. \ref{fig: prompt}.

As shown in Fig. \ref{fig: prompt}, our prompt consists of two types of prompts: the prefix prompt and the input prompt. The prefix prompt is used to illustrate the basic reasoning rules, and the input prompt provides perceptual information. The prefix prompt introduces task settings, input and output format specifications, and reasoning sub-tasks. Specifically, the task settings aim to inform the LLM of its role as a core controller, while the format specifications serve to standardize the LLM's output, facilitating post-processing tasks such as parsing the viewpoint selection from the LLM's output. The reasoning sub-tasks aim to assist the agent in decomposing the task into several sub-tasks. However, there are slight variations in the sub-tasks between the CoT mining phase and the navigation inference phase, as shown in Figure \ref{fig: prompt}. For the REVERIE task, the agent is not only required to navigate to the target location, but also to identify the target object. Consequently, sub-tasks such as referring expression and further exploration become necessary. Conversely, in CoT mining, as decision-making is provided, there is no requirement for additional reasoning regarding referring expression or exploration. Instead, the focus lies on explaining the rationale behind the decisions.

\subsection{Action prediction} \label{predict}
The LLM receives the prompt as input and produces predictions. Among these output items, viewpoint selection is critical. The selected viewpoint corresponds to one of the viewpoints stored in the Memory Topological Map, allowing the shortest-path planner $F_S$ to compute a path to the target viewpoint.

%% file: sections/setup.tex
\section{Experimental setup}

\subsection{Datasets} 
\textbf{REVERIE} dataset comprises 21,702 instructions, with an average instruction length of 18 words. The agent is required not only to navigate to the target location but also to point out the referred object by selecting the appropriate object bounding box.


\textbf{R2R} dataset contains 21,567 instructions with an average length of 29 words. R2R is a well-established VLN benchmark, which features sequential instructions without object localization requirements.

For the R2R dataset, we follow the experimental settings of previous works such as NavGPT \cite{zhou2023navgpt} and MapGPT \cite{chen2024mapgpt}, conducting experiments solely on a subset of Val Unseen split of the R2R dataset. This setting serves two purposes: 1) considering the API costs, running the entire VLN dataset is impractical; 2) since neither NavGPT, MapGPT, nor our method involves training, any data split (validation seen, validation unseen, or test) remains unseen relative to the navigator. For the REVERIE dataset, we conduct evaluations on the entire Val Unseen split. To the best of our knowledge, our approach is the first LLM-core method to conduct evaluations on the \textbf{entire} val unseen dataset for the REVERIE dataset.

\subsection{Evaluation Metrics} 
We utilize the pre-defined evaluation metrics of the VLN task. (1) Trajectory Length (TL): the agent's path length in meters. (2) Success Rate (SR): the agent's success rate of stopping at a viewpoint where the referred object can be observed. (3) Oracle Success Rate (OSR): the agent's success rate of having a viewpoint along the trajectory where the referred object can be observed. (4) Success rate weighted by Path Length (SPL): the agent's navigation success rate weighted by the length of the trajectory length. (5) Navigation Error (NE): the agent's mean navigation error in meters. (6) Remote Grounding Success rate (RGS): the agent's success rate of pointing out the referred object. (7) Remote Grounding Success rate weighted by navigation Path Length (RGSPL): the agent's success rate of pointing out the referred object weighted by the trajectory length.

\subsection{Implementation Details}
We utilize the ChatGPT (gpt-3.5-turbo-1106 with 16K context window) and GPT-4o as our LLMs. Note that we have also experimented with Gemini \cite{team2023gemini} and Llama2 \cite{touvron2023llama}, both of which did not perform as well as GPT-3.5 and GPT-4o. Therefore, we choose to present the results with GPT-3.5 and GPT-4o. To understand the visual content of the viewpoint's images, we use the Faster-RCNN \cite{girshick2015fast,anderson2018bottom} pre-trained on the Visual Genome \cite{krishna2017visual} dataset to detect the surrounding objects and use the InstructBLIP-13B \cite{li-etal-2023-lavis} as our VLM to caption the images of viewpoints. 

%% file: sections/result.tex
\section{Results}
\subsection{Main Results}

We initially evaluate our method against several learning-based approaches on the REVERIE benchmark. We categorize existing methods into three groups: Training, Pre-training, and No train. Most VLN methods are trained from scratch and belong to the Training category. More recent approaches are pre-trained on several proxy tasks and fine-tuned on downstream VLN tasks, falling into the Pre-training category. In contrast, our LLM-based method {\em requires neither fine-tuning nor pre-training}.

As shown in Tab. \ref{tab:reverie}, our method yields significant improvements over methods in the Training category (which only fine-tune on the VLN datasets) across all metrics, demonstrating its superior capability in few-shot scenarios. However, compared to methods in the Pre-training category, our approach does show certain performance gaps. Pre-training methods typically involve a two-step process of pre-training followed by fine-tuning on VLN datasets. This process incurs high training overhead and often faces challenges in convergence. Additionally, these methods can suffer from limited generalization to novel environments. In contrast, our proposed MC-GPT leverages a versatile LLM, allowing for rapid adaptation to VLN tasks without the need for extensive additional training. Moreover, it enhances the model's ability to generalize across different datasets and environments, making it a robust and efficient solution for varied applications in VLN tasks.


In Tab. \ref{tab:r2r}, we evaluate MC-GPT on 72 various scenes of the R2R dataset. By using GPT-3.5, our MC-GPT surpasses NavGPT \cite{zhou2023navgpt} and MapGPT \cite{chen2024mapgpt} across all metrics. The comparison demonstrates our advantages in memory construction and navigation strategy: on one hand, compared to NavGPT which uses an additional LLM to generate navigation memories, our memory map is more concise, has less noise, and does not require calling LLM. On the other hand, since the MapGPT is also equipped with a map, the advantage of our method suggests that the use of a navigation chain of thoughts is indeed helpful. We also conduct experiments with GPT-4o, and the comparison is shown in Tab. \ref{tab:r2r}. The results indicate that employing GPT-4o significantly enhances the navigator's performance compared to using the GPT-3.5 model. Compared to the contemporary work MapGPT, our method, when using GPT-4o, shows slightly lower performance in Success Rate (SR) but surpasses MapGPT in both Oracle Success Rate (OSR) and Navigation Error (NE). This indicates that using the most advanced LLM can improve performance and complement the design presented in our paper. In other words, our method can demonstrate its effectiveness across different LLMs.


\begin{table}[h]
  \centering
  \caption{Performance comparison on REVERIE.}
  \resizebox{1.0\columnwidth}{!}{
    \begin{tabular}{c|l|llllll}
    \toprule
    \multicolumn{1}{c|}{\multirow{2}{*}{Settings}} &
   \multicolumn{1}{c|}{\multirow{2}{*}{Methods}}  &      \multicolumn{6}{c}{Validation Unseen} \\
\cmidrule{3-8} & & SR↑   & OSR↑  & SPL↑  & TL    & RGS↑  & RGSPL↑ \\
    \midrule
    \multicolumn{1}{c|}{\multirow{3}{*}{Pre-training}}
    & RecBERT \cite{hong2021vln} & {30.67}  & {35.02}  & {24.90}  & 16.78 & {18.77}  & {15.27} \\
    & Airbert \cite{guhur2021airbert}  & 27.89 & 34.51 & 21.88 & 18.71 & 18.23 & 14.18 \\
    & DUET \cite{chen2022think}  & 46.98 & 51.07 & 33.73 & 22.11 & 32.15 & 23.03 \\
    \midrule
     \multicolumn{1}{c|}{\multirow{6}{*}{Training}}
     &Random & 1.76  & 11.93 & 1.01  & 10.76 & 0.96  & 0.56\\
     &Seq2Seq-SF \cite{anderson2018vision} & 4.20   & 8.07  & 2.84  & 11.07 & 2.16  & 1.63 \\
     &RCM \cite{wang2019reinforced} & 9.29  & 14.23 & 6.97  & 11.98 & 4.89  & 3.89 \\
    &SMNA \cite{self} & 8.15  & 11.28 & 6.44  & 9.07  & 4.54  & 3.61 \\
     &FAST-Short \cite{ke2019tactical} & 10.08 & 20.48 & 6.17  & 29.7  & 6.24  & 3.97\\
     &FAST-MATTN \cite{qi2020reverie} & 14.4  & {28.20}  & 7.19  & 45.28 & 7.84  & 4.67 \\
    \midrule
    \rowcolor{gray!20}
    No Train & MC-GPT (Ours) & 19.43 & 30.25 & 9.65 & 24.50 & 8.86 & 5.14 \\
    \bottomrule 
    \end{tabular}%
    }
  \label{tab:reverie}%
\end{table}%

\begin{table}[h]
  \centering
  \caption{Performance comparison on 72 various scenes of the R2R dataset. ``Exp\#'' refers to the number of GPT experts and ``Dist'' the utilization of distance information.}
    \begin{tabular}{llcl|rrr}
    \toprule
    Methods & LLMs  & \multicolumn{1}{l}{Exp\#} & Dist  & \multicolumn{1}{l}{SR↑} & \multicolumn{1}{l}{OSR↑} & \multicolumn{1}{l}{NE↓} \\
    \midrule
    NavGPT \cite{zhou2023navgpt} & GPT-3.5 & 2     & w/    & 16.7  & 26.4  & 8.02 \\
    MapGPT \cite{chen2024mapgpt} & GPT-3.5 & 1     & w/o   & 19.4  & 29.6  & 8.48 \\
    \rowcolor{gray!20}
    MC-GPT (Ours)  & GPT-3.5 & 1     & w/o   & \textbf{22.1} & \textbf{34.2} & \textbf{7.76} \\
    \midrule
    MapGPT \cite{chen2024mapgpt} & GPT-4 & 1 & w/o &\textbf{41.2} &61.6 &5.80 \\
    \rowcolor{gray!20}
    MC-GPT (Ours) & GPT-4o & 1 & w/o &32.1 & \textbf{68.8} & \textbf{5.42} \\
    \bottomrule
    \end{tabular}%
  \label{tab:r2r}%
\end{table}%

\begin{figure*}[h]
    \centering
    \includegraphics[width=1.0\textwidth]{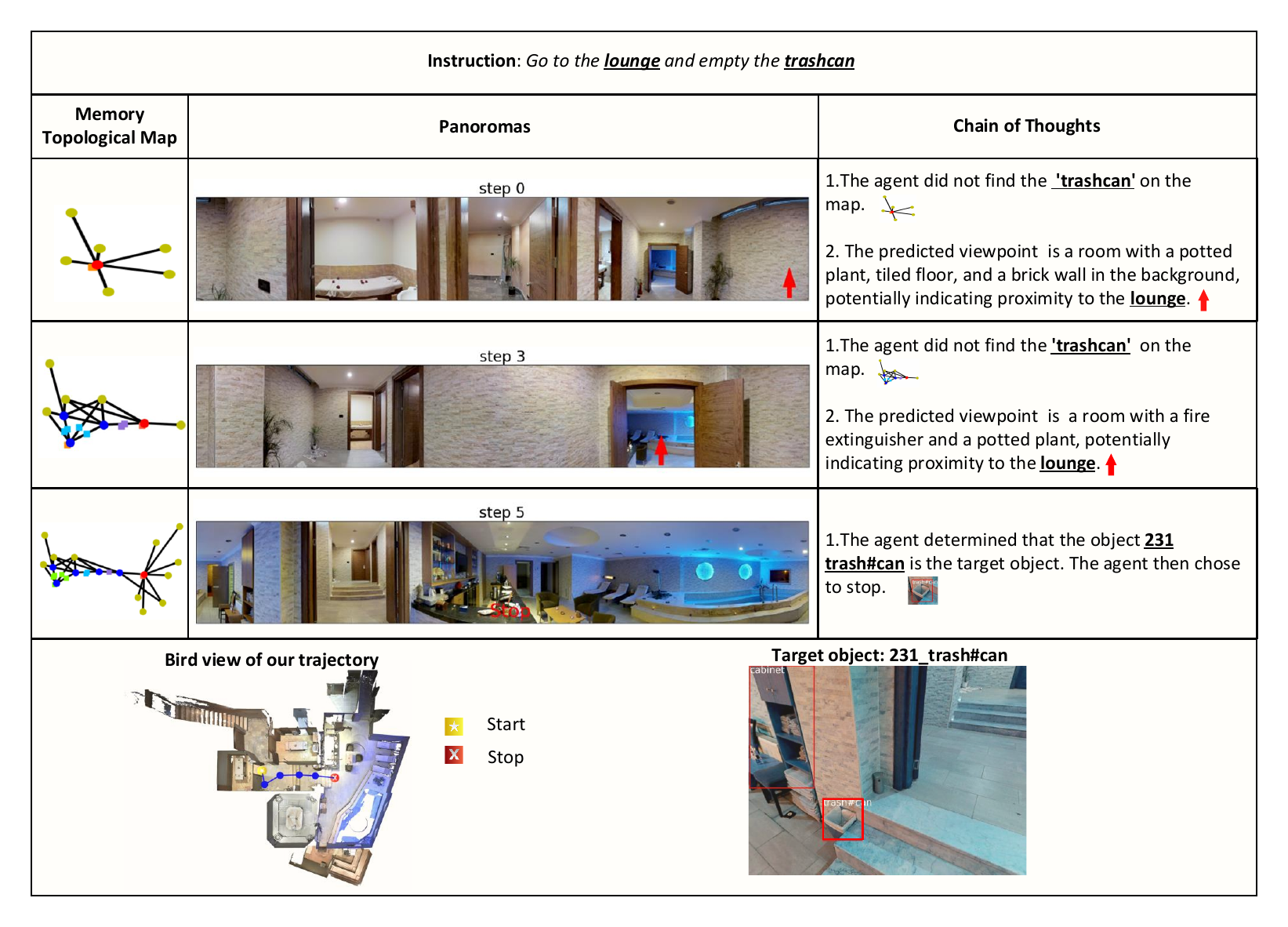}
    \caption{Visualization of a successful navigation task. Initially, the agent is presented with multiple navigable viewpoints, including a stairway leading to the upper floor and several doors. Utilizing the proposed Memory Topological Map and Navigation Chain of Thoughts, the agent steadily advances toward the lounge area and finally identifies the target object -- the trash can.}
    \label{fig: visualization}
\end{figure*}

\subsection{Ablation Study}
In this section, we conduct an ablation study on individual components. We randomly select 100 instructions from the REVERIE Validation Unseen dataset and establish a subset for experiments.

\textbf{Is the navigation chain of thoughts essential?} Our navigation chain of thoughts is included in a demonstration example, so we illustrate the necessity of the navigation thought chain by comparing the ablation effects of the demonstration example. As shown in Tab. \ref{tab:demo}, when no example is provided, that is, the LLM's prompt only contains basic rules (prefix prompt), the results indicate that the LLM possesses commendable abilities in rule comprehension and navigation reasoning. Then we consider whether providing random demonstration examples is helpful: we compare two methods for querying a demonstration example, one based on instruction similarity, and the other based on destination room type similarity. Surprisingly, when using the former measure, the navigation performance is similar to that without using examples. This may be due to the presence of many function words in the instructions, leading to significant differences between the matched examples and the test scenes, and making the examples less useful as references. However, when using destination room type as a similarity measure, the performance is significantly improved, indicating that examples with similar destination room types are more meaningful references. Note that a demonstration example should contain complete navigation steps, otherwise they are not very helpful for the agent. In short, an appropriate demonstration example with a complete chain of thoughts can guide the agent for better navigation.


\textbf{Does it help if the demonstration example is close to the test instance?} As shown in Tab. \ref{tab:rt}, we conduct a comparative experiment: under the same conditions of using destination room type similarity for example querying, the closer the similarity, the more similar the scene between the demonstration example and the test instance. As shown in Tab. \ref{tab:rt}, when the destination room type of the example is the ground truth (RT-GT), the most suitable demonstration example can be matched, which is more helpful for navigation. When using Python's regular expressions to extract room type from instructions (RT-RE), there is a slight decrease in performance, but it is within an acceptable range. In order to be closer to reality and reduce prediction costs, we adopt the latter method.

\begin{table}[h]
  \centering
  \caption{Comparison of various settings of Navigation Chain of Thoughts on REVERIE. ``Demo'' refers to the utilization of a demonstration example and ``Query-sim'' refers to the similarity measure of the querying. ``Step'' refers to the number of steps of a demonstration example.}
    \begin{tabular}{ccc|llll}
    \toprule
    Demo  & Query-sim & Step & SR↑   & OSR↑  & SPL↑  & TL \\
    \midrule
     $\times$ & $\times$ & $\times$    & 18.00    & 30.00    & 8.83  & 20.13 \\
     $\surd$     & Instruction & All   & 17.00    & 29.00    & 9.78  & 21.78 \\
     $\surd$     & Room type & 1 & 19.00    & 34.00    & 8.73  & 22.12 \\
     \rowcolor{gray!20}
     $\surd$     & Room type & All   & \textbf{23.00}    & \textbf{34.00}    & \textbf{11.16} & 23.80 \\
    \bottomrule
    \end{tabular}%
  \label{tab:demo}%
\end{table}%

\begin{table}[h]
  \centering
  \caption{Effect of  accuracy of destination room type prediction on navigation performance. ``RT-GT'' refers to the ground-truth room type, and ``RT-RE'' refers to the room type predicted by using Python's regular expressions.}
    \begin{tabular}{l|llllll}
    \toprule
    Methods & SR↑   & OSR↑  & SPL↑  & TL    & RGS↑ & RGSPL↑ \\
    \midrule
    RT-GT & \textbf{25.00}    & \textbf{39.00}    & 11.14 & 22.99 & 7.00     & 4.07 \\
    \rowcolor{gray!20}
    RT-RE & 23.00    & 34.00    & \textbf{11.16} & 23.80  & \textbf{9.00}     & \textbf{4.22} \\
    \bottomrule
    \end{tabular}%
  \label{tab:rt}%
\end{table}%

\begin{figure*}[htbp]
    \centering
    \includegraphics[width=1.0\textwidth]{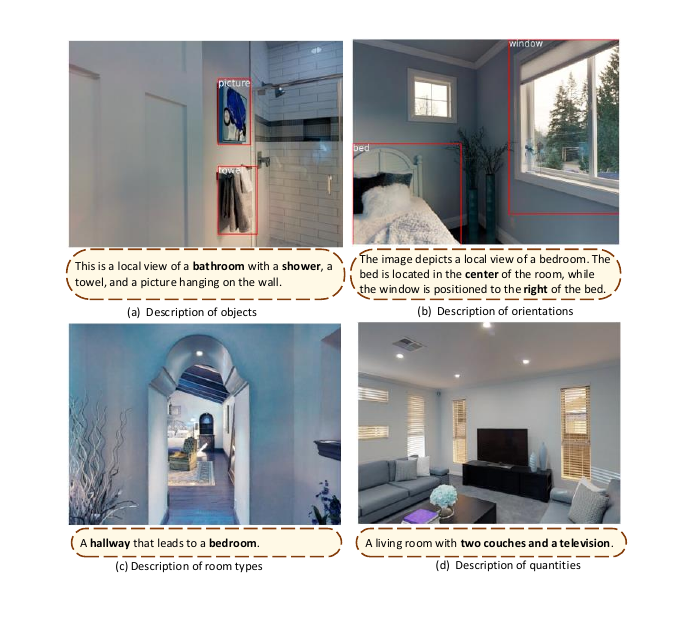}
    \caption{Examples showcasing the conversion from visual observations to textual descriptions. InstructBLIP exhibits enhanced description capabilities leveraging our object detection context (a). Furthermore, in specific scenarios, it can articulate object orientations (b), identify room categories (c), and employ quantifiers (d) for descriptive purposes.}
  \label{fig:observation_appendix}
  \vspace{-1.0em}
\end{figure*}

\subsection{Case Study}
In Fig. \ref{fig: visualization}, we present a visualization of a successful navigation example. As the number of navigation steps increases, the scale of the Memory Topological Map gradually expands. This map serves not only to record the distribution of viewpoints, aiding the navigator in determining the shortest path, but also to clearly illustrate the relationships between viewpoints and objects. This functionality enables the LLM to make semantic navigation predictions. Throughout each step of the chain of thoughts, it is evident that the LLM primarily engages in reasoning about two pivotal questions: ``Is the target object present?'' and ``What serves as the basis for predicting the navigation point?'', which aligns with our reasoning sub-tasks settings as depicted in Fig.\ref{fig: prompt}. This illustrates the LLM's thorough utilization of the sub-task settings and perceptual information provided within our prompt.

\subsection{Cases of image captioning} \label{blip}
In Section \ref{map}, we use InstructBLIP-13B \cite{li-etal-2023-lavis} for generating captions from observations. In this section, we present the specific prompt used with InstructBLIP-13B and provide several illustrative examples. The size of the input image is 640×480 pixels, containing the results of object detection. When objects are detected in the image, we prompt the InstructBLIP with the object detection results: \emph{``Context: This is a local view of a room with several objects including [detected objects]. Provide a brief description of the image".} When no objects are detected in the image, we simply prompt the InstructBLIP: \emph{``Provide a brief description of the image"}. 

In Fig. \ref{fig:observation_appendix}, we present some examples of our transformation from visual observations to textual descriptions. We find that InstructBLIP can better describe observations using our provided object detection results and, in some cases, can also describe object orientations, determine room types, and use quantifiers for descriptions. However, it is inevitable for InstructBLIP to generate illusions, which is highly related to the working mode of generative models. Therefore, relying solely on InstructBLIP's perception is insufficient for the navigator. This is also one of the motivations behind proposing the memory topological map in this paper: to reduce noise and improve accuracy.

%% file: sections/conclusion.tex
\section{Conclusion}
Our work aims to address the challenges in LLM-based VLN navigators by proposing novel approaches to memory construction, navigation strategy diversity, and integration of navigation components. We provide a solution for retaining spatial relationships and facilitating navigation decisions. Additionally, our Navigation Chain of Thoughts module leverages human navigation examples to enhance the diversity of navigation strategies, contributing to more robust and flexible navigation systems. Experimental results on the REVERIE and R2R datasets confirm the effectiveness of our approach in enhancing interpretability and bridging the gap between human and robotic navigation. Moving forward, further research is needed to explore additional enhancements and optimizations to fully leverage the potential of LLMs in empowering VLN navigators.